# An Agent-based Modeling Framework for Sociotechnical Simulation of Water Distribution Contamination Events

M. Ehsan Shafiee and Emily M. Zechman

North Carolina State University

**Abstract.** In the event that a bacteriological or chemical toxin is introduced to a water distribution network, a large population of consumers may become exposed to the contaminant. A contamination event may be poorly predictable dynamic process due to the interactions of consumers and utility managers during an event. Consumers that become aware of a threat may select protective actions that change their water demands from typical demand patterns, and new hydraulic conditions can arise that differ from conditions that are predicted when demands are considered as exogenous inputs. Consequently, the movement of the contaminant plume in the pipe network may shift from its expected trajectory. A sociotechnical model is developed here to integrate agent-based models of consumers with an engineering water distribution system model and capture the dynamics between consumer behaviors and the water distribution system for predicting contaminant transport and public exposure. Consumers are simulated as agents with behaviors defined for water use activities, mobility, word-of-mouth communication, and demand reduction, based on a set of rules representing an agents autonomy and reaction to health impacts, the environment, and the actions of other agents. As consumers decrease their water use, the demand exerted on the water distribution system is updated; as the flow directions and volumes shift in response, the location of the contaminant plume is updated and the amount of contaminant consumed by each agent is calculated. The framework is tested through simulating realistic contamination scenarios for a virtual city and water distribution system.

## 1 Introduction

Drinking water distribution systems are critical infrastructure systems. Because they provide potable water to a large population of consumers from centralized sources, they create vulnerability in a community to both accidental pathogen outbreaks and intentional attacks (1). For example, accidental introduction of pathogens may occur due to pump failure, pipe breaks, and polluted source water, and intentional attacks may be initiated at tanks, treatment plants, and exposed water mains (2). Historical outbreaks due to contaminated water have caused severe public health consequences, including hospitalization and death of vulnerable segments of the population (3).

The public health consequences of an outbreak or attack do not depend on the characteristics of the contaminant intrusion alone. Instead, the decisions and behaviors of human actors as they interact with the pipe network can influence the number of exposed consumers and the propagation of a contaminant plume. The U.S. Government Accountability Office (1) recommended that system vulnerability may be reduced by changing human actions and interactions, including strengthening the communication between actors who distribute information during an event and training utility operators for outbreak mitigation. Woo and Vicente (4) and Vicente and Christoffersen (5) have described the water contamination outbreak as a sociotechnical system, which is a system that is characterized by strong interactions between social and technical factors that govern the emergent system performance. Optimization of a technical system without consideration of social interactions tends to degrade the performance and increases the unpredictability of the system. Glouberman (6) investigated an outbreak in Walkerton, Ontario, and emphasizes that the occurrence and effects of a water contamination event cannot be attributed to one culprit alone. The study suggests that attempts to strengthen a system by focusing on one component alone may ignore events and interactions that can have important stabilizing or destabilizing effects. While the analysis of Walkerton focused on the interactions that increase the vulnerability of the system and that influence detection of the event, the interactions between human actors and the infrastruc- ture system as the event unfolds can also influence the transport of contaminant in the network and determine the emergent public health consequences. For ex- ample, as a contaminant spreads through a system, consumers may become sick and stop consuming water or change their water use due to warnings from public health officials or their peers. At the same time, a water utility may take action to mitigate the event after the detection of the contaminant. The utility may use both outreach activities, such as broadcasting boil water orders, and operational procedures, such as closing valves and opening hydrants. The actions taken by consumers and decision-makers may cause hydraulic conditions in the network to fluctuate outside of expected ranges and, therefore, the contaminant plume to deviate from a predicted propagation. The unpredictability introduced through consumer interactions may create difficulties for utility managers in identifying the most effective responses plans to protect public health.

Simulation studies typically consider water distribution pipe networks in isolation and neglect dynamic interactions among the contaminant in the pipe network and consumer demand decisions. A Complex Adaptive System (CAS) approach is developed here for the water distribuiton contamination event. A CAS is characterized by a set of interacting agents that influence emergent system properties through dynamic feedback loops (7; 8; 9; 10). Agent-based modeling (ABM) is a computational model for simulating the actions and interactions of autonomous agents in a CAS to evaluate the collective effect on system properties (11). Agents are modeled to receive information about their environment, have goals, and select actions to change the environment and meet goals. Additionally, an agent can receive information from other agents and interact with them. The ABM approach has been applied in water supply management in limited contexts to explore decision-making strategies for increasing water supply capacity and to simulate consumers and their reactions to water pricing (12; 13; 14; 15; 16). Preliminary research has explored simulation of water distribution contamination through coupling water distribution system simulation with ABM, and this approach was applied to evaluate the public health consequences of contamination events for a small virtual city of 5000 residents (17; 18; 19). The study presented here makes a new contribution to threat management research by comparing public health consequences that are predicted through the ABM methodology to results that are obtained from an engineering model alone. This research demonstrates and quantifies the change in the predicted hydraulic conditions in the network that the decentralized decisions of consumers can produce. While utility managers and their reactions to mitigate the consequences of water events can significantly impact the consequences of a contamination event, this study explores consumer behaviors, and utility actions are not included for simulation in the present study. The framework developed and presented here introduces advances in behavioral simulation beyond the previous studies. Improvements to the simulation of human behavior include incorporation of established models for a word-of-mouth mechanism, simulation of a demographically heterogeneous population, improved simulation of exposure, and utilization of real data for simulating the expected protective actions of consumers during contamination events. This study applies the ABM approach for a virtual case study, which is a municipality of 150,000 residents. The case study is realistic and includes a large population of consumers and a complex water distribution system. A new metric is used to evaluate the location of a contaminant plume in a network over a time series, and results show that for large doses of potent chemicals, the hydraulics in a network are reversed in flow directions from normal operating conditions. In preliminary studies, ABM was proposed and implemented as an approach to address water contamination events; this study emphasizes the importance of ABM for studying water events as sociotechnical systems by comparing it with the models that are traditionally used for analysis.

## 2 Agent-Based Modeling Framework

A set of actors, including end-use consumers, water utility managers, public officials, the mass media, and public health agencies, interact through sharing information when a contaminant is introduced in a water distribution system, as illustrated in Fig. 1. Individual actors and organizations choose from a set of potential actions, based on their interpretations of the event, and these actions can directly or indirectly change hydraulics in the pipe network. For example, water utility managers may first become aware of contamination through unusual water quality data that is collected by an early warning sensor system or through consumer complaints. Water utility managers may implement operational strategies to control the propagation of the contaminant plume in the water distribution system. Public officials may alert residents about potential or confirmed contamination by implementing, for example, a boil water notice. Consumers may comply with notices and warnings by implementing protective actions, such as avoiding contact with tap water, and alerting peers about a threat. As a result of these interactions, hydraulic conditions in the water distribution system may fluctuate due to direct operations, such as opening hydrants and closing values, and due to reductions in consumer demands.

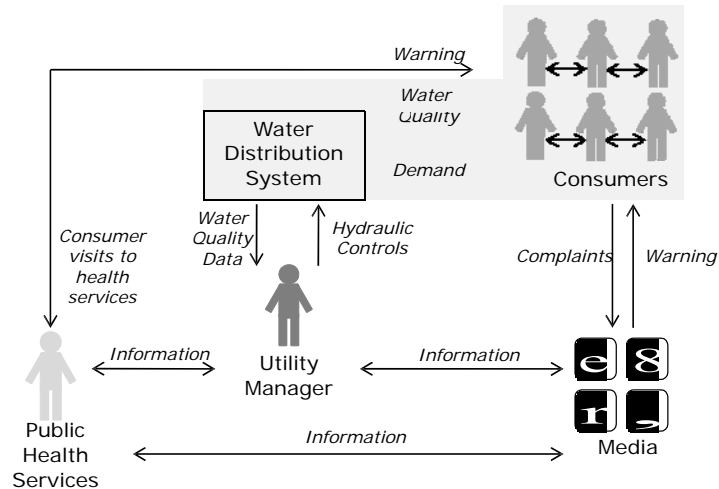

Fig. 1. Interactions among actors and the water distribution system in a water distribution system contamination event. The study described here captures the interactions among consumers and the water distribution model, as shown by the shaded box.

A modeling framework is described here to capture the interactions among consumers and the water distribution system by coupling a water distribution simulator EPANET (20) with an ABM system, AnyLogic (21). Consumers are represented as individual agents, who adapt their behavior based on the information they receive from the water distribution system and other agents. When the simulation begins, consumer demands are represented as aggregated demands that are exerted on the water distribution system at nodes, and EPANET calculates the flow volumes, flow directions, and water quality within the pipe network. The simulation proceeds at discrete time steps for hydraulic calculations in EPANET, and at each time step, water quality information is passed from the water distribution system model to the agents. Consumers are simulated to ingest water, and once a consumer has accumulated a critical dose of the contam-inant, the consumer is flagged as exposed and responds to exposure by passing information to other consumers and reducing water usage. Changes in demands are translated to the water distribution simulation, and the hydraulic conditions for successive time steps are calculated by EPANET. In this way, the feedback loop between the consumers and the water distribution system is established.

The procedure that consumers use to select behaviors is represented through a set of rules, which include if-then relationships and probabilistic functions. The behaviors that are included in the simulation are the timing and volume of water ingestion; changes in water usage through protective action strategies; travel among nodes; and word-of-mouth communication among consumers. Agent attributes and rules of behavior are described in the following sections.

## 2.1 Demographic information

Agents are initialized with diverse characteristics to represent a heterogeneous population. Data is used from a study that reports statistics for age, gender, and weight, for the U.S. population, as grouped into 11 discrete age groups (22) (Fig. 2).

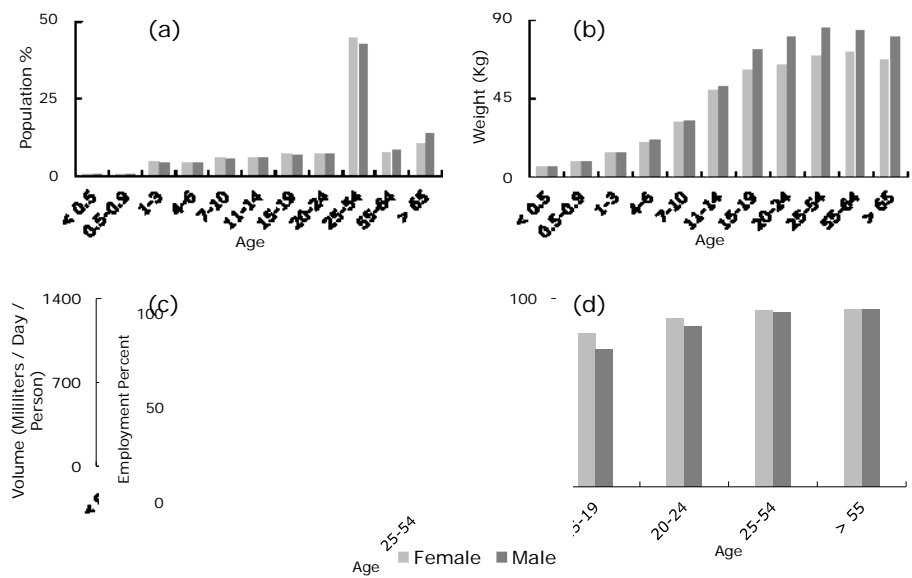

Fig. 2. Demographic information for the U.S. population shown as the average for an age group (a) Age (22) (b) Weight (22) (c) Volume of ingested tap water (22) (d) Employment (23).

## 2.2 Volume and Timing of Daily Ingestion

Each agent is initialized with a specific pattern for using and consuming water, based on gender and age group, using the distribution shown in Fig. 2c. Two approaches, a fixed approach and a probabilistic approach, are used and compared to simulate the volume of water that each agent consumes each day. The fixed approach simulates that each agent ingests the expected value of the distribution, or 0.93 L. The probabilistic approach uses an exponential distribution to assign random volumes to different consumers based on the mean value for each age group. A value for the daily consumption volume is assigned to each consumer using Eqn. 1, which is the inverse form of an exponential distribution function (24):

$$v = -v_m \ln(1 - p) \quad (1)$$

where $v$ is the volume of water, $v_m$ is the mean volume associated with each age group in Fig. 2, and $p$ is a probability that is randomly generated between zero and one.

A consumer agent is assigned five times during a day when it consumes tap water. The fixed approach divides the volume of water that each consumer drinks uniformly among five times: 7:00, 9:30, 12:00, 15:00, and 18:00 (25). The probabilistic approach was developed to specify probability density functions for the times at which a consumer takes three daily meals, depending on the timing of previous meals (Fig. 3) (26). Minor meals are taken at the mid-point between major meals to simulate five daily ingestions.

As agents ingest contaminated water, the mass of contaminant in an agent's body accumulates. Once an agent has ingested a critical dose, he is considered exposed. The critical dose varies for different contaminants.

## 2.3 Water Demand Behaviors and Self-Protective Strategies

Simulation of water end-use behaviors is based on data collected by the American Water Works Association (2008). This study reports that 70% of the total residential water demand for U.S. households is used for indoor activities, and these indoor water end-uses are grouped into six categories based on the ultimate appliance: washing clothes (15.4% total-water use), toilet (18.5%), shower (11.6%), faucet (11.2%), leakage (9.8%), and other miscellaneous indoor uses (3.5%) (27). Consumer agents are simulated with the ability to suspend water use for four of these indoor activities, including washing clothes, showering, using the faucet, and miscellaneous indoor uses, which compose 41.7% of a household's total water use.

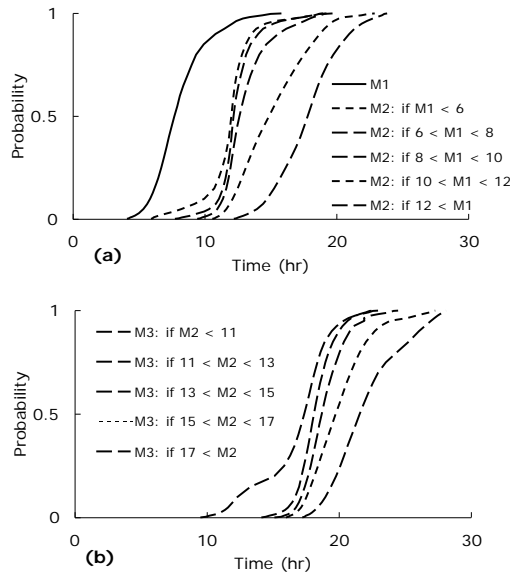

Fig. 3. Cumulative distribution functions (CDF) represent the times at which three major meals are taken. M1, M2, and M3 are the times for Meals 1, 2, and 3, respectively. (a) CDF for M1 and M2, (b) CDF for M3.

Once consumers become exposed to a contaminant or are alerted by peers about a threat, they may adjust their demand decisions and change typical consumption of tap water for indoor uses. Results of a survey about expected behaviors in a water contamination event are used to better represent how consumers make decisions about reducing water use during a contamination event. The survey was conducted to explore how respondents may change nine water activities related to indoor use in response to information about water contamination (28). Respondents were asked to quantify how likely continuing certain water use activities would be to endanger their health when drinking water has been contaminated. Survey results (Fig. 4) were used to calculate the probability that a consumer agent will continue each activity, and each water activity corresponds to one of the four water end-use groups listed above (washing clothes, showering, using the faucet, and miscellaneous indoor uses). When a consumer agent selects to alter its water use behavior, the probability of suspending each use is evaluated independently, and the total reduction in total water demand is computed.

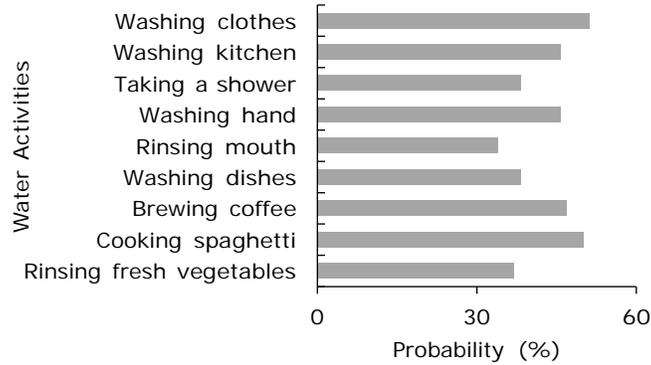

Fig. 4. Probabilities for suspending water activities once a consumer is alerted about an event, based on results reported by (28)

As agents change their behaviors, the base demand of each hydraulic node is updated at each time step:

$$bd^0_t = \frac{\sum_{i=1}^{i=K}(1-RF_i)}{K} \times bd_t \qquad (2)$$

where $bd_t$ is the original base demand at a node at time step t; K is the number of consumers located at the node at time step t; $RF_i$ is the reduction factor decision made by consumer i at the node; and $bd^0_t$ is the new base demand at the node. The parameter $RF_i$ varies between 0.035 and 0.417 to represent the reduction in demand corresponding to the end uses that are discontinued for an agent.

2.4 Mobility

During a contamination event, consumers may move across the boundary of a contaminant plume as they travel to work, places of business, or residences, and may become exposed to the contaminant by drinking water at any of these locations. The percentage of employed adults, distinguished by age group and gender, was reported by the Bureau of Labor Statistics for cities in the U.S. (23) (Fig. 2d) and is used in simulating mobility. To simulate the movement of consumers in a municipality, each agent is assigned a residential node, a non-residential node, the time at which the agent leaves its residential node, the length of time spent traveling, and the time it leaves the non-residential node to return to its residential node. Employed agents spend approximately eight hours at non-residential nodes, and unemployed agents visit commercial nodes or remain at residential nodes during a day. The travel time between two nodes is based on the Euclidean distance and is subtracted from the time a consumer spends at a residential node.

Data describing demand patterns and the time series of the population at each node are used to establish mobility patterns that are consistent with the water distribution model. Demand patterns that specify the daily time series of nodal demand are obtained from the input data for the hydraulic model. These patterns are normalized and multiplied by the maximum population at a node to derive the number of consumers that should be at a node at each time step, and these numbers are used to assign mobility parameters to agents.

### 2.5 Word-of-Mouth Communication

During hazardous events, individuals may receive information through many varied channels, and word-of-mouth communication can significantly affect the behavior of a population (29). Victims may identify their own unsafe actions and encourage family members, friends, peers, and colleagues to discontinue water use activities. Once consumers experience exposure symptoms, they may adopt protective behaviors and warn others about the danger.

While a few models exist for simulating communication among peers (e.g., 30), a cluster model was developed for simulating communication specifically during an emergency event (31) and was selected for implementation within the ABM framework. The cluster approach is similar to a small world network model (32), but the cluster approach specifies a unidirectional flow of information and specifies relatively less communication among agents. Each agent is specified as an information isolate or a member of a cluster. In a cluster, individuals are assigned one of several roles, including an original source, intermediate members, and ultimate receivers. Original sources are typically individuals who are informed about current events, communicate with many individuals, and have a strong influence in their immediate community. A warning message can originate from the original source or the intermediate actors in each cluster (Fig. 5). The time it takes to pass each message is assumed for this study as one step of the hydraulic simulation (typically 15 minutes to 1 hour). Upon receiving a message, agents in a cluster wait one time step before passing the message to the next receiver.

## 3 Illustrative Case Study: Mesopolis

Mesopolis is a virtual city that was developed for threat management research (33) and is used to demonstrate application of the ABM framework for a realistic case study. The city is modeled with an historical development, beginning in the 1800s as a harbor town and growing to include residential, commercial, and industrial areas, in addition to a naval base, an airport, and a university (Fig. 6). A river that flows through the city from south to north provides the main source of fresh water. Water is withdrawn from the river at a location 13 miles south of the city and pumped through a main pipe to the southern edge of Mesopolis, where the pipe branches to deliver water to an East Treatment Plant and a West Treatment Plant. The West Treatment Plant is an older plant and provides the majority of the water demand for the west side of the city, which is separated from the east side by the river. The East Treatment Plant supplies water for the eastern side of the city, and, in addition, for a large portion of the western side during peak demands. A skeletonized hydraulic simulation model represents the network using 1588 nodes, 2058 pipes, one reservoir, 13 tanks, and 65 pumps. Four different demand patterns are specified, which include residential, commercial, industrial, and naval demands. Commercial nodes include churches, schools, grocery stores and malls. Demands at industrial nodes are typically constant through a 24-hour period, corresponding to three 8-hour shifts, while the demands at other types of nodes increase during the day and reduce to nearly zero at night. Based on the demands simulated in Mesopolis, the population is calculated as 146,716 persons, and the distribution of the population among node types is shown in Table 1. The mobility patterns are established using the demand patterns and population distribution among nodes (Fig. 7). For the word-of-mouth simulation, the cluster size is set at 15, with one information isolate, one original source, two intermediate receivers, and 11 ultimate receivers.

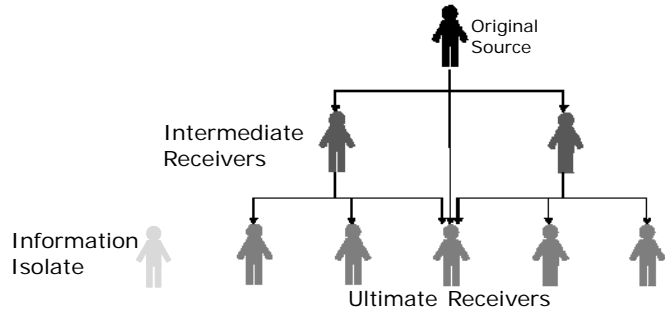

Fig. 5. Framework for sending and receiving messages within a cluster of individuals (31). Arrows show the direction of messages passed from senders to receivers.

## 4 Contamination Events

A large number of potential intentional attacks and accidental outbreaks can threaten the water quality in a distribution system. A set of both biological and chemical contaminants are explored, including E. coli, Norwalk-like virus, and arsenic. The characteristics of the contamination events are based on a risk assessment study, which reports the location and timing for the worst case events that could occur in Mesopolis and the maximum number of exposed consumers that would be expected for each event (34; 35). The risk assessment study was conducted to evaluate the severity of event without considering consumer actions or reductions in demands. A hydraulic water distribution simulation model was coupled with an engineering model with an optimization methodology to identify worst-case threats, which are characterized by the time that the contaminant is injected, the duration over which the contaminant was released, and the hydraulic demand multiplier (the hydraulic demand multiplier represents the fluctuation of demands with seasons). To estimate loading values for the bacterial contamination, probability distributions were created to describe the likelihood of any load of bacteria injected to a system, based on literature that describes bacterial outbreaks. Due to a lack of occurrence and documentation about wide-spread arsenic contamination, two loading events were created to represent low-impact and high-impact cases. A selected number of events that were identified through the risk assessment are used as simulation scenarios for this study (Table 2), including intrusions of E. coli, the Norwalk-like virus, 75 kg of arsenic, and 300 kg of arsenic at the East Treatment Plant and West Treatment Plant, for a total of eight contamination events.

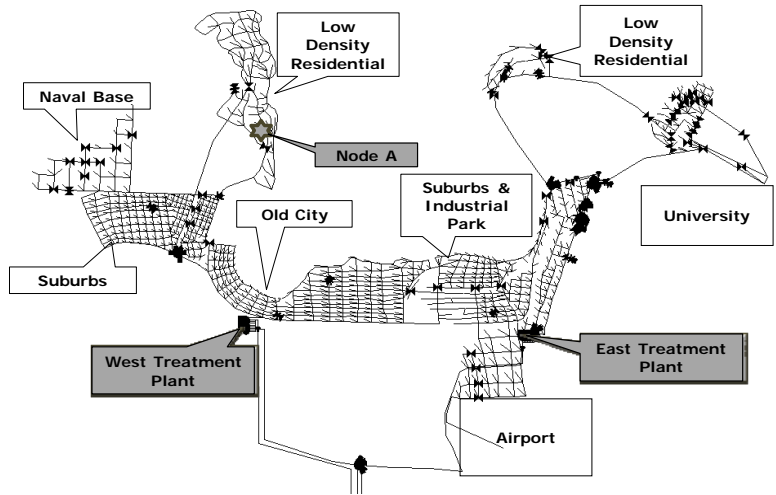

Fig. 6. Schematic of the Mesopolis water distribution system

Table 1. Distribution of the Mesopolis population among different types of agents

| Types of Consumers | Population |
| --- | --- |
| Employed consumers | 86,776 |
| Employed consumers at commercial nodes | 26,034 |
| Students | 19,457 |
| Employed consumers at industrial nodes (over 3 shifts) | 21,186 |
| Unemployed consumers who visit commerical nodes | 18,552 |
| Unemployed consumers remaining at residential nodes | 6,755 |

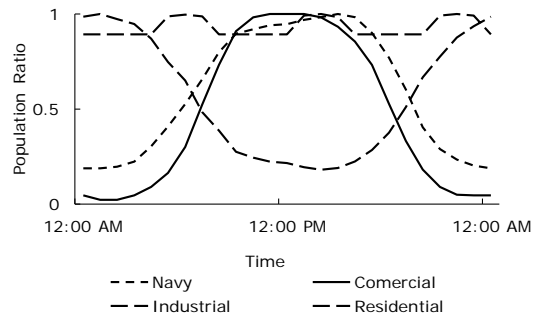

Fig. 7. Percentage of consumers at demand node types. The percentage is based on the aggregated maximum population for each type of demand node and demand patterns.

For each of the eight contamination events, the value for the critical dose for each contaminant is calculated using exposure information and models available in the literature. The critical dose represents the dose at which a person experiences symptoms. The critical dose (infectious dose) for E. coli and Norwalk-like are nine and 15 cells, respectively (36). The arsenic critical dose varies based on the weight of a victim (37):

$$d_c = 5.0 \times 10^{-8} w_m \qquad (3)$$

where $d_c$ is the arsenic critical dose in kilograms, and $w_m$ is the consumer's body weight in kilograms. For the initial set of simulations, all consumers are assigned a weight of 70 kg, with a corresponding arsenic critical dose of 3.5 mg.

## 5 Modeling Scenario

Five modeling scenarios are constructed to represent increasing levels of complexity in the model (Table 3). Model 1 represents the static engineering model, where all consumers drink water at the same time each day and of the same volume (0.93 L), remain at residential nodes without moving, are exposed after ingesting 3.5 mg of arsenic, and do not adapt their demands or communicate with other consumers. Model 2 includes probabilistic simulation of the volume and timing of consumer tap water ingestion, though consumers still remain at residential nodes and maintain water consumption. Model 3 incorporates daily mobility of consumers in the network. Model 4 includes a feedback loop between the consumers and the network, where consumers decrease demands based on exposure, and hydraulic conditions are altered dynamically. Model 5 includes the adaptive behaviors of Model 4 and, in addition, consumer agents communicate and receive warnings through the word-of-mouth mechanism to update their demands.

Table 2. Accidental pathogen outbreaks and intentional attacks used for contamination events, adopted from (35). Seasons include Fall/Winter (F/W) and Winter (W).

| Event Type | Location | Pathogen or Toxic Chemical | Demand Multiplier (Season) | Injection Starting Time | Injection Ending Time | Contaminant Load |
|---|---|---|---|---|---|---|
| Accidental | West Treatment Plant | E. coli | 0.85 (F/W) | Day 1, 8am | Day 3, 7pm | 36M doses |
| | | Norwalk-like | 0.90 (F/W) | Day 1, 8am | Day 4, 6pm | 38M doses |
| | East Treatment Plant | E. coli | 0.80 (F/W) | Day 1, 12am | Day 2, 1am | 74M doses |
| | | Norwalk-like | 0.95 (F/W) | Day 1, 12pm | Day 3, 12pm | 91M doses |
| Intentional | West Treatment Plant | Arsenic | 0.60 (W) | Day 1, 12pm | Day 2, 2pm | 75 kg |
| | | Arsenic | 0.60 (W) | Day 1, 6pm | Day 2, 12am | 300 kg |
| | East Treatment Plant | Arsenic | 0.60 (W) | Day 1, 6pm | Day 1, 7pm | 75 kg |
| | | Arsenic | 0.65 (W) | Day 1, 7pm | Day 1, 12am | 300 kg |

Table 3. Modeling scenarios for the ABM framework

| Model Mechanisms | Model Static → Dynamic | | | | |
|---|---|---|---|---|---|
| | 1 | 2 | 3 | 4 | 5 |
| Ingesting Volume and Times | Det. | Prob. | Prob. | Prob. | Prob. |
| Mobility | No | No | Yes | Yes | Yes |
| Adaptation of Consumers | No | No | No | Yes | Yes |
| Word-Of-Mouth Communication | No | No | No | No | Yes |

# 6 Results

## 6.1 Total Exposure

For each modeling scenario, the simulation duration has been set to 8 and 10 days, according to the type of event, to provide a baseline for comparing the modeling scenarios. The models described here do not simulate the responses and actions of the utility. Without the interactions of the utility, such as opening hydrants to flush contaminated water, consumers continue to use water until a contaminant has been completely consumed, which occurs after 8 and 10 days for the intentional and accidental events, respectively. As shown in the following results, new dynamics arise in the simulation after 6 days due to the adaptations of the population. Future and on-going research will explore the interaction of utility actions during a contamination event (38).

Each of the five models was executed for 10 random trials, for a total of 50 simulations for each one of the eight contamination events. Fig. 8 shows the results for the contamination events (E. Coli outbreak, Norwalk-like virus outbreak, 75-kg arsenic event, and 300-kg arsenic event) at the West and East Treatment Plant. Results are presented as the average number of exposed consumers, and the error bars show the range of the number of exposed consumers over 10 trials. The stochasticity that is introduced by the probability distributions in the modeling of the behavior of agents results in a small variation in the predicted number of exposures. For all models and contamination scenarios, the range of exposed consumers varied within a range of 300-600. Compared to the total number of exposed consumers, the range of variation is small due to the size of the contamination events, which introduce a large load of the contaminant. In addition, for many simulations, the contaminant remains for a significant amount of time at many nodes, which gives the consumers at those nodes repeated opportunities to consume contaminated water. Though there is variation in mobility and ingestion, over time, similar numbers of consumers are exposed across the set of 10 realizations for each unique combination of contamination event and model.

The static model, Model 1, predicts the highest number of exposed consumers across all models for the contamination events at the East Treatment Plant. For each of the four events, Model 2 predicts between 40-50% the number of exposed consumers as predicted by Model 1. Model 2 includes stochasticity in the volume of water ingested by each consumer and in the timing of consuming water. As a result, the range of the contaminant mass ingested by each consumer is wider, and while some consumers ingest much more of the contaminant, fewer consumers ingest contaminant above the critical dose. Mobility (included in Model 3) decreases the predicted number of exposed consumers by a small percentage, as more consumers travel away from contaminated nodes and do not ingest a critical dose. Model 4 includes adaptive behavior, and because consumers change their water demand once they have become exposed, there is more contaminant that remains in the network for a longer time period, and additional consumers are exposed. Model 5 shows a small decrease in the predicted number of exposed consumers, as consumers who are warned through the word-of-mouth mechanism before they become exposed are protected from the contaminant.

The results for events at the West Treatment Plant show different patterns for the predicted number of exposed consumers when compared to events at the East Treatment Plant. Model 1 predicts the highest number of exposed consumers for all events except the 300-kg arsenic event. Model 2 shows a decrease (40 to 65% reduction compared to Model 1) in the predicted number of exposed consumers, but mobility (simulated in Model 3) increases the number of consumers who are exposed. This is because for events initiated at the West Treatment Plant, the contaminant is confined to a small part of the network in larger concentrations, so that the most significant impact of mobility is that the number of individuals from the eastern side of the city visit the contaminated area during the day and ingest contaminated water, becoming exposed. Including the adaptive nature of consumers in Model 4 increases the number of exposed consumers, as the contaminant lingers in the system without being consumed by exposed consumers. The word-of-mouth mechanism, included in Model 5, protects a small percentage of the total population, which is seen when compared to Model 4.

The 300-kg arsenic event at the West Treatment Plant is significantly different in predicted consequences. Stochasticity in drinking patterns (Model 2) does not decrease the number of exposed consumers to the same extent as other events. Mobility increases the number of exposed consumers above the number predicted by Model 1, producing a greater increase than for any of the other events or models. The adaptive behavior of consumers as they reduce their demands increases the number of exposed consumers further. The impact of the word-of-mouth communication is greatest for this event, and reduces the number of exposed consumers by approximately 10,000. These results are explained by the high contaminant load of 300 kg. This event produces contaminant in concentrations above 1.5 $\frac{mg}{L}$ lingering for 19 hours at 67% of the terminal nodes in the western part of the city. Most consumers at nodes on the western side of the city consume a critical dose of arsenic due to the high concentration in the water. The number of the residential consumers in the western portion of the city is 37,463, and the number of individuals predicted as exposed using Model 3 is an average of 38,865, indicating that mobility results in the exposure of more consumers though they do not reside in the western part of the city.

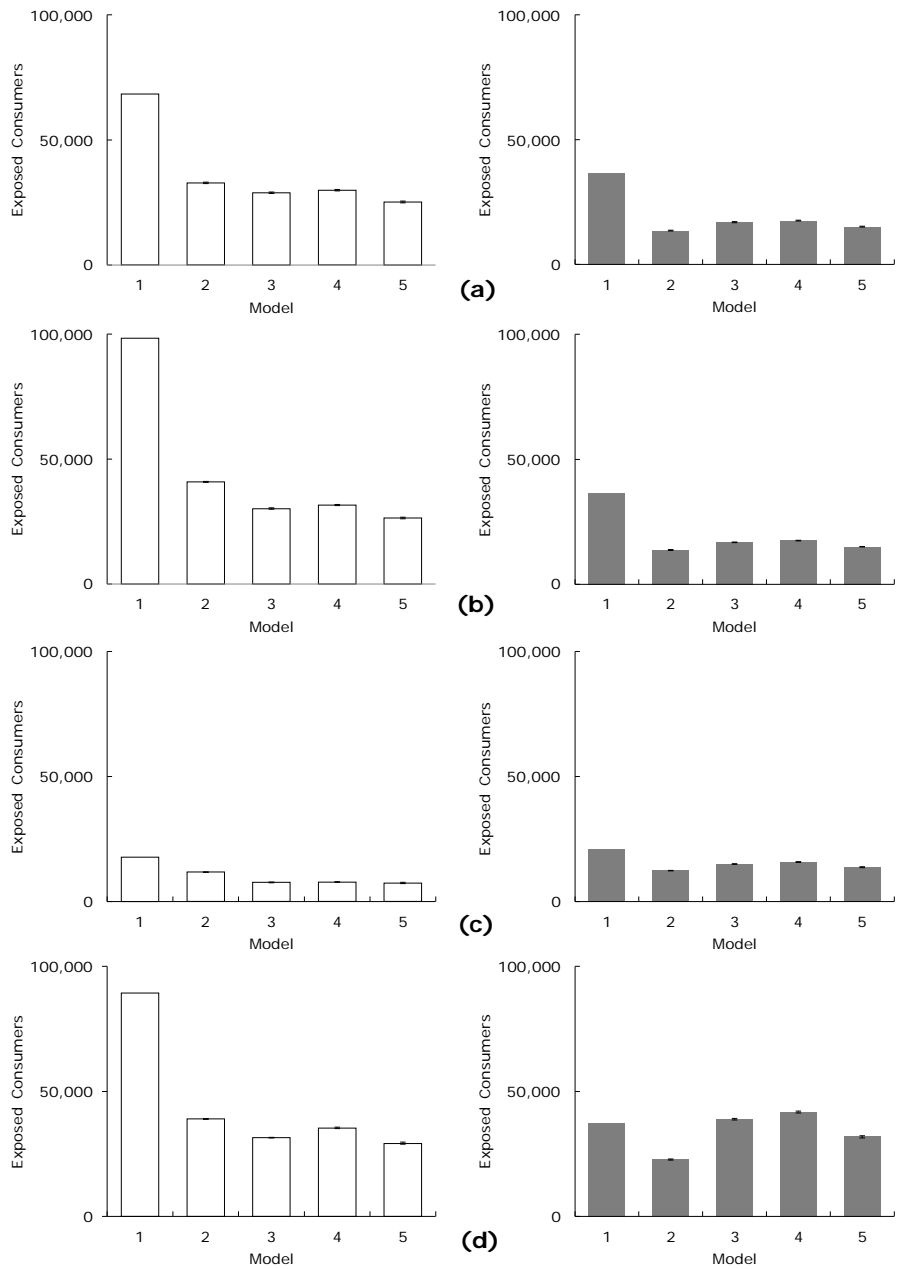

Fig. 8. Average number of exposed consumers due to contamination events at the East Treatment Plant (left column, white bars) and West Treatment Plant (right column, dark bars) over 10 trials for each model. Error bars show the range of exposed consumers over 10 trials. Contaminant Events are (a) E.coli, (b) Norwalk-like, (c) 75-kg arsenic, (d) 300-kg arsenic.

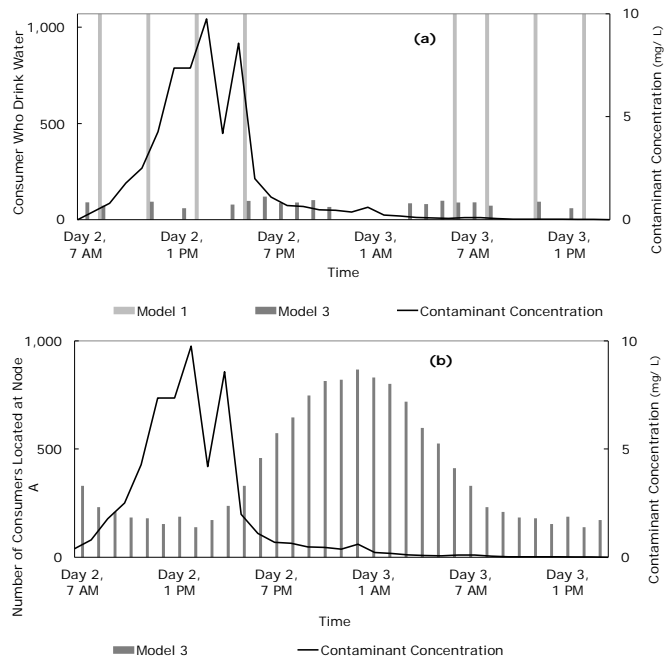

Fig. 9. Fig. 9. Results for one simulation of the 75-kg arsenic event at the West Treatment Plant. (a) Contaminant concentration profile at Node A and the timing of consumer ingestions for Model 1 and Model 3 (b) Contaminant concentration profile at Node A and the number of consumers located at the node, simulated using Model 3.

In all of the events at both treatment plants, there is a notable decrease in predicted exposed consumers between Models 1 and 2. The decrease when stochasticity is introduced to consumer behavior is heightened for these events because the contamination events were originally designed through optimization to have the most impact for the deterministic model, Model 1 (35). The impacts of optimizing events for simplified modeling is further explored through Fig. 9. Fig. 9a shows the profile of contaminant concentration at Node A when 75 kg of arsenic is introduced at the West Treatment Plant (the location of Node A is shown in Fig. 6). The timing for consumer consumption is simulated using Models 1 and 3, and the bars show the average over 10 simulations for each model. Model 3 represents the ingestion timing and volume for Models 3, 4, and 5, because stochasticity in consumer water activities and mobility are included in these models. Model 1 predicts that all 1071 people located at Node A drink water on the second day at 12:00 P.M., 3:00 P.M., and 6:00 P.M., when the contaminant is at high concentrations at that location. Model 3 predicts a more uniform distribution of ingestions at Node A. Model 1 predicts that the consumers at Node A drink a total of 796 liters of water, and Model 3 predicts 598 liters of water are ingested during the 34-hour window shown in Fig. 9a.

Fig. 9b demonstrates the mobility of consumers, which is included in Models 3, 4, and 5. Consumers move throughout the city, and as a result, there are only a few hundred consumers at Node A when contaminant concentrations are high. As a result, less than 300 consumers drink water at Node A when the contaminant concentration is at the highest value. For the 10 simulations of each model, the volume of ingested water at Node A is an average of 262 liters (standard deviation of 29 liters) for Models 3 and 4, and an average of 237 liters (standard deviation of 20 liters) for Model 5. These volumes are much lower than the volume of water that is predicted to be ingested at Node A using Model 1. These results demonstrate that a worst-case scenario that is designed using vulnerability analyses and a static model may not be the worst-case scenario when the complexities of adaptations and interactions in an event are taken into account.

## 6.2 Dynamics of Exposure

Fig. 10 shows the dynamics of consumer exposure for four of the contamination events, including 300-kg arsenic events and Norwalk-like outbreaks at the West Treatment plant and the East Treatment Plant, as predicted by all five Models. For each event, Model 1 shows a stepwise behavior in the increasing number of exposed consumers, which is due to the uniformity of consumer behaviors. All consumers in the population drink at the same five events during one day, and each consumer ingests the same volume of water, though the amount of contaminant that is ingested varies due to diverse contaminant concentrations throughout the pipe network. For Model 1, consumers are counted as exposed only after each of the five daily drinking events, leading to the stepwise behavior. For the remaining models, there is stochasticity in the timing of ingestion events, leading to a more continuous increase in the number of exposed consumers. In simulating the consequences of the Norwalk-like event (Figs. 10c and 10d), Models 2-5 predict that some consumers are exposed to the critical dose while the contaminant is still being released, in contrast to Model 1, which predicts that consumers become exposed only once the entire load has been injected into the system. This is due to the variability in the amount of ingested water among consumers in Models 2-5; some consumers drink large enough quantities to become exposed much earlier in the event.

Fig. 11 shows for these same events the time series of consumers who are exposed, consumers who are warned through word-of-mouth communication, and consumers who change their water use, simulated using only Model 5. The line representing the number of consumers who change their water use generally follows the exposed consumers; as more consumers are exposed, more consumers are warned. In addition, consumers who are warned also warn others. The number of consumers who change their water use does not exceed 70,000 for any of the simulations due to the limitations imposed by the word-of-mouth mechanism. When an agent becomes exposed, the warning goes to agents in the same cluster, but does not spread to other clusters.

6.3 Effects of Adaptive Consumer Behaviors on System Hydraulics

The hypothesis of this work is that as consumers change behaviors in response to the contaminant, the hydraulic conditions in the network are altered. To illustrate differences among the models in the predicted movement of the contaminant, the Coincident Population Plume Index (CPP) is introduced here. The CPP represents the coincidence, or coinciding location and timing, of the population and the contaminant plume. The CPP assesses the fraction of consumers who are located at nodes where the contaminant concentration is greater than zero:

$$CPP(t) = \frac{\sum_{i=1}^{i=n} \frac{p_i(t)}{1-\delta[Q_i(t)]}}{P} \quad (4)$$

where $CPP(t)$ is the Coincident Population Plume Index at time step $t$ of a simulation; $p_i(t)$ is the population at node $i$ and time step $t$; $n$ is the total number of terminal nodes in a water distribution system; $Q_i(t)$ is the contaminant concentration at node $i$ at time $t$; $\delta(x)$ is the Dirac delta function (39); and $P$ is total number of consumers in the model. The function $\delta[Q_i(t)]$ takes on a value of infinity when the contaminant concentration is zero, and a value of zero when the contaminant concentration is greater than zero.

The CPP index is computed for the E. coli outbreak at the East Treatment Plant (Fig. 12a). The E. coli outbreak is selected because the duration of the contaminant intrusion is shorter than the Norwalk-like virus event and shows dynamics of the $CPP$ clearly. Models 1 and 2 predict the similar behaviors for the $CPP$ index, and the $CCP$ value drops off during days 6-8, as the contaminant leaves the system. Diurnal patterns are present during days 6-8, due to the filling and draining of a tank, which removes the contaminant from the pipes during the night and re-introduces the contaminant to the pipes as the tank drains during the day. For the first six days of the simulated event, Models 3, 4, and 5 show similar values for $CPP$, indicating that the adaptive behaviors of the consumers do not influence the location of the contaminant plume. Later in the event, during days 6-8, the location of the plume varies widely among these models. Specifically, Model 3 shows lower $CPP$ values, as residents continue to consume water and the contaminant, and contaminant leaves the system. For Model 4, some consumers change their demands so that the contaminant lingers in the system longer, and for Model 5, as additional consumers change demands, the contaminant stays in the system and covers a larger portion of the network.

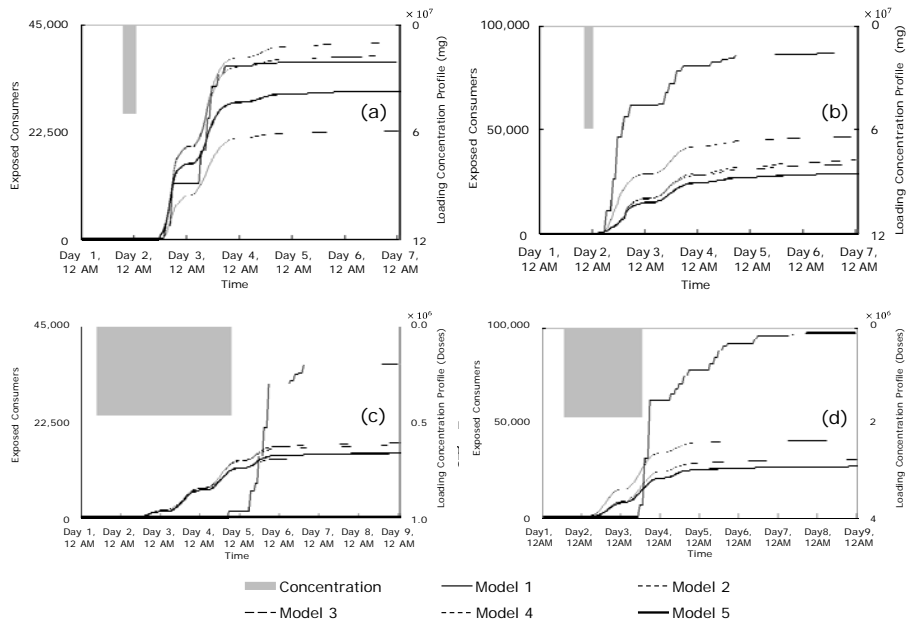

Fig. 10. Time series of the average number of the exposed consumers over 10 trials using Models 1-5. Contaminant loading profile is shown as shaded area. (a) 300-kg arsenic event at the West Treatment Plant, (b) 300-kg arsenic event at the East Treatment Plant, (c) Norwalk-like at the West Treatment Plant, and (d) Norwalk-like at the East Treatment Plant.

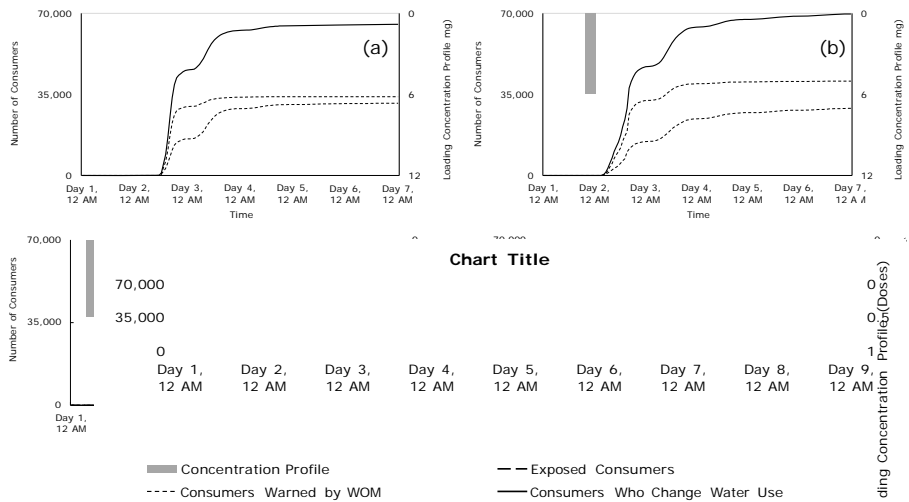

Fig. 11. Time series of the average (over 10 trials) number of the exposed consumers, consumers who are warned, and consumers who change their water use, simulated using Model 5. (a) 300-kg arsenic event at the West Treatment Plant, (b) 300-kg arsenic event at the East Treatment Plant, (c) Norwalk-like at the West Treatment Plant, and (d) Norwalk-like at the East Treatment Plant.

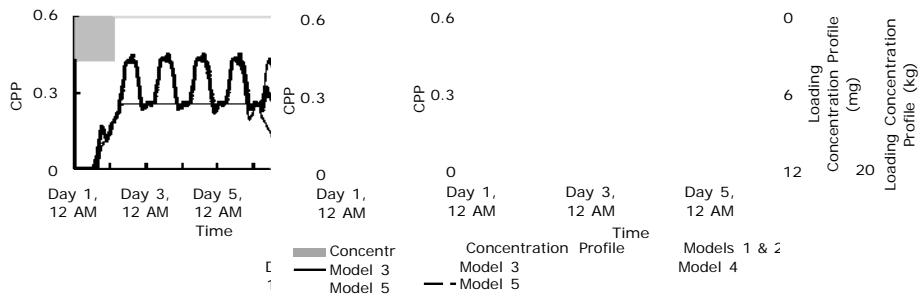

Fig. 12. Time series of Coincidence Population Plume Index for a) E.coli outbreak at East Treatment Plant and b) 300-kg arsenic event at West Treatment Plant simulated using Models 1-5.

The $CPP$ index is also computed for the 300-kg arsenic event at the West Treatment Plant (Fig. 12b). Models 1 and 2 (plotted as one line in Fig. 12) show the same behavior for the value of $CPP$, as the location of consumers and the movement of the plume is identical between the two models. The number of consumers at contaminated nodes does not change for Models 1 and 2 over the first six days of the event (Fig. 12b). The behavior of the $CPP$ for Model 3 shows an oscillating pattern, due to the daily movement of consumers, as many travel to the western side of the city during the daytime, and back to residential nodes on the eastern side of the city during the nighttime. The mobility patterns among Models 3, 4, and 5, are the same, but the behavior of the $CPP$ index for Models 4 and 5 depart from that of Model 3. For Models 4 and 5, the $CPP$ is lower than the $CPP$ for Model 3. The adaptive behaviors of the consumers (changes in demands for Model 4, and changes in demands in addition to the word-of-mouth mechanism for Model 5) have influenced the water system to the extent that the predicted location of the plume shifts. As shown in Fig. 13, and described in the following paragraph, Model 5 predicts that the plume moves to the central part of the city, where nodes are non-residential and fewer consumers are present, during the later part of the simulation when demands have been adapted. Model 5 predicts a lower value for $CPP$ during the last few days of the simulation.

The 300-kg arsenic event at the West Treatment Plant demonstrates strong dynamic behaviors that lead to dramatic changes in the predicted contaminant plume, as consumers are exposed early in the event and adapt their demands quickly. The hydraulics in the system is impacted to such an extent that the normal direction of flow in the system is reversed. Fig. 13 depicts the total spread of the contaminant plume, as predicted by Models 1 and 5. Under Model 1, the contaminant is constrained to the western side of the city. Under normal conditions, water flows from east to west to meet peak demands; using Model 5, hydraulic conditions are changed to such an extent that water flows from west to east due to the number of consumers that have reduced their water demand, and the contaminant plume reaches a greater number of nodes. This mechanism breaks what has been identified as a hydraulic barrier, or a division in the pipe network over which water does not flow or flows uni-directionally under normal operating conditions. This case demonstrates the utility of the ABM approach to identify unexpected emergent dynamics that may occur due to adaptive behaviors.

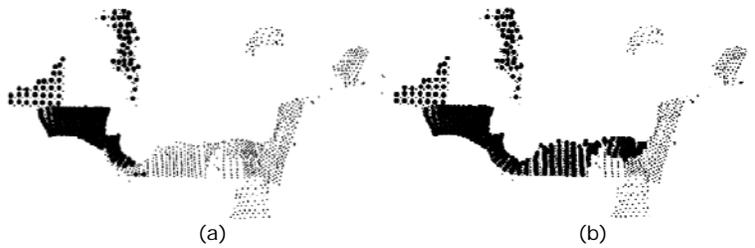

(a)  (b)

Fig. 13. 300-kg arsenic event introduced at the West Treatment Plant as predicted by (a) Model 1 and (b) Model 5. Contaminated terminal nodes indicated by large dark circles. Terminal nodes that remain clean throughout the event are shown as small dark circles.

## 7 Sensitivity Analysis

In each of the five Models, a few assumptions were made and can impact the simulation results. For example, consumers are simulated to recognize and respond quickly to symptoms of exposure and to notify peers immediately. Zechman (40) conducted a sensitivity analysis of these parameters for a small water network. Further studies are needed to tune these behaviors to represent realistic behaviors. A sensitivity analysis is conducted here to evaluate the impact of two modeling parameters, critical dose and the word-of-mouth framework, on public health consequences.

### 7.1 Critical Dose for Arsenic

The initial modeling assumed that all consumers have a body weight of 70 kg, and as a result, all consumers respond to a critical dose of 3.5 mg (or 0.05 $\frac{mg}{kg}$ body weight) of arsenic. There may be considerable uncertainty in the size of arsenic dose that causes symptoms to appear for any individual. Three additional cases are considered here, where consumers respond to critical doses of arsenic of 0.035, 0.050, and 0.071 $\frac{mg}{kg}$ body weight, and consumer agents are initialized with weights that are generated probabilistically to better represent a heterogeneous population. Critical doses represent the upper bound, lower bound, and median values of exposure (37). The age group and gender of each agent (initialization of these parameters is described above) translates to a mean value for the weight of a consumer based on the U.S. representative statistical distribution (shown in Fig. 2). To generate the weight of each consumer, an exponential distribution function is used (Eq. 5) and ensures that for a large sample, the average mean of the weight of the sample matches the average for the original data set:

$$w = -w_m \ln(1 - p) \qquad (5)$$

where $w$ is a weight, $w_m$ is the mean weight of each age group in Fig. 2, and $p$ is a probability that is randomly generated between zero and one. Fig. 14 shows the predicted number of exposed consumers for varying critical doses. As expected, a higher critical dose results in a lower predicted number of exposed consumers. For Model 1, these modeling variations result in significant differences in the results. For Models 2-5 at the West Treatment Plant for the 300-kg arsenic event (Fig. 14b), using the homogeneous population under-predicts the number of exposed consumers, even when compared to results corresponding to a critical dose of 0.0710 $\frac{mg}{kg}$ body weight. For events at the East Treatment Plant, simulating a homogeneous population matches more closely the heterogeneous population with a critical dose of 0.05 $\frac{mg}{kg}$ body weight. The dynamics of the more extreme events (Figs. 14b and d) exacerbates the unpredictability of model results.

## 7.2 Word-of-Mouth Framework

The modeling structure that is adopted for the word-of-mouth mechanism may also significantly impact the predicted number of exposed consumers. Parameters in the word-of-mouth framework that can be investigated include the cluster size and the number of intermediate members. Analysis demonstrated that varying the number of intermediate members does not impact the number of exposed consumers in the range of two to eight, when the size of the cluster is kept constant at 15. Varying the size of the cluster, however, can significantly alter the results (Fig. 15). As the number of members in a cluster increases, more individuals will be informed of an event and change their water use, resulting in fewer exposed consumers. The number of exposed consumers is reduced by 50% as the cluster size increases from 10 to 30, indicating that results may be significantly sensitive to the word-of-mouth simulation. The parameters that best represent the community structure and communication characteristics of a population should be identified to facilitate a better understanding of the dynamics that may occur in an event.

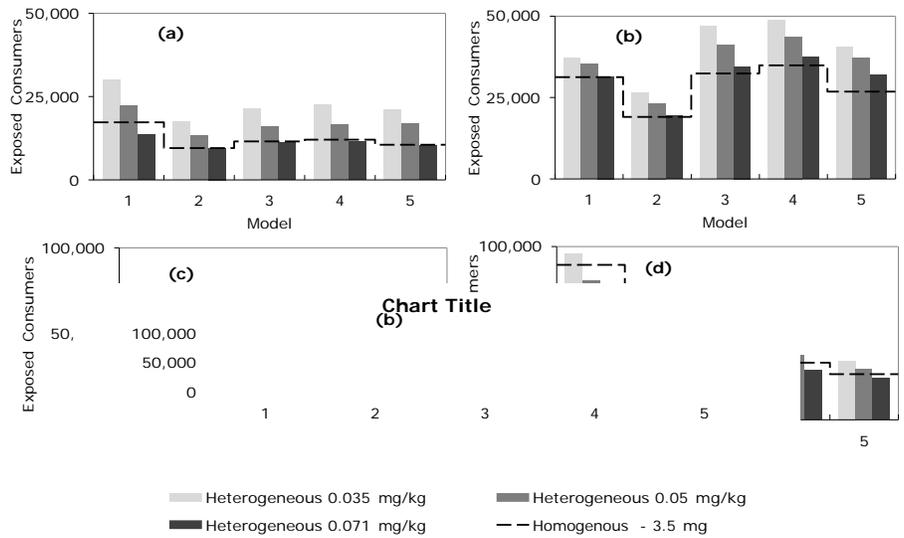

Fig. 14. Consumer response curves for the heterogeneous population simulated with various critical doses compared to the base case of a homogeneous population. Simulated for (a) 75-kg arsenic event at the West Treatment Plant; (b) 300-kg arsenic event at the West Treatment Plant; (c) 75-kg arsenic event at the East Treatment Plant; and (d) 300-kg arsenic event at the East Treatment Plant.

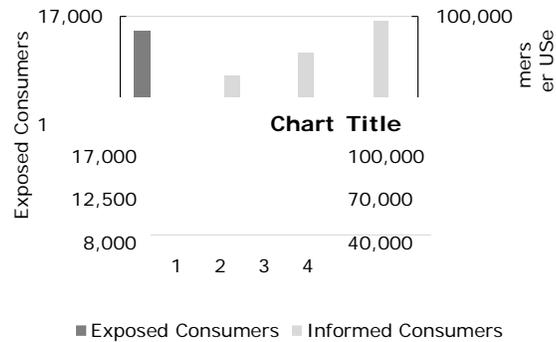

Fig. 15. Number of consumers who change their water use and the number of exposed consumers for varying cluster sizes.

# 8 Discussion and Conclusions

An ABM framework is described and demonstrated for simulating contamination of water distribution systems. Sociotechnical simulation integrates modeling of consumer behavior and a hydraulic model of the pipe network to predict the number of exposed consumers for an event. The ABM simulates consumers as agents with mechanisms for mobility, word-of-mouth communication, probabilistic estimations of the volume of water ingestion, and probabilistic timing for water ingestion. The research presented here makes a new contribution to research investigating the significance of consumers adapting their demands in event of a contamination through a comparison of the results of the sociotechnical model with an engineering and exposure model alone (34; 35). Analysis is conducted here to create new understanding about the simulation of a dynamic event: results demonstrate the spatial and temporal variation in consumer demands and in the movement of the contaminant plume as consumer activities of increasing complexity (e.g., mobility and communication) are included in the simulation. New results present here demonstrate that in some cases, adaptations of consumer demands can alter the predicted hydraulics of the system, the movement of the contaminant plume, and as a result, public health consequences. Specifically, for potent events that are isolated in the western portion of the city, the fluctuation of the hydraulics changes the flow directions in the network from normal operations and breaks a hydraulic barrier, which leads to a higher number of exposures. The simulation approach that is developed here can benefit utility managers and public health officials in developing plans for mitigating consequences of contamination events. Specifically, managers can gain insights to the potential dynamics that can influence the direction and flows of water in the system. Vulnerability analysis that is conducted by considering an engineering model in isolation may mis-identify the worst-case scenarios for contaminant intrusion, and response plans that are developed without consideration of adaptive behaviors may disregard important dynamics that influence the performance of selected protective actions.

This study also develops a new metric for evaluating the movement of the plume and the population within the network. The Coincident Population Plume index is defined to represent the movement of polluted water to nodes that are highly populated, and it can be used to evaluate the change in the plume due to adaptations of consumer behaviors. The index gives a concise metric for displaying the change in the vulnerable population based on additional complexity in the model. For this study, values of the index illustrate that models of increasing complexity predict that the polluted water is available to a wider range of consumers. By including accurate representations of the complex sociotechnical interactions, more conservative predictions of consumer exposure can be obtained. The results of this modeling framework can lead to better understanding of the impact of events on public health and to better selection of components that should be hardened and mitigation strategies.

This study also improved the simulation of consumer behavior beyond what has been implemented previously to provide a more realistic representation of the potential behaviors of individuals. Much of this behavioral simulation is still rudimentary, however, especially regarding the reactions of consumers to exposure and information about a threat. For example, the simulation assumes that consumers respond within one hour of becoming exposed or receiving a warning from a peer. Previous work demonstrated that the emergent consumer exposure is sensitive to timing information (40). New information is needed from social science studies about the timing and risk-aversion of consumer reactions to hazards, sickness due to specific contaminants, and public warnings, to better conceptualize and parameterize the model.

Finally, this research conducted a sensitivity analysis to explore the influence of the social model on the predicted consequences. The analysis demonstrated that the cluster modeling assumptions and settings could significantly influence the predicted results. New research for modeling and parameterizing social networks and their functions in hazards can be incorporated, as it becomes available, in the ABM framework. Beyond the research demonstrated here, the modeling can be enhanced through inclusion of information about consumer response to different types of media and warning messages, and consumer response to symptoms. The dissemination of warning notices by utility managers and how they influence the adoption of protective actions are not part of this modeling study, but are investigated in on-going work (41).

The sociotechnical simulation developed here forms a bridge to connect research conducted in behavioral science with engineering management. ABM provides an approach to incorporate probabilistic information about consumer choices with hydraulic simulation. Additional research can investigate the extent to which social networks and engineering preparedness determine the resilience of a community to hazardous events. Additional agents can be defined to represent utility managers, media, and public health officials to capture additional dynamics due to human behavior and errors. An extended framework can be developed and is being investigated to evaluate the effectiveness of alternative response actions that can be selected by public officials, as response actions can include a wide range of hydraulic and social response, such as opening hydrants and closing pipes (19); and warning consumers through media and emergency siren vehicles (42).

# 9 Acknowledgments

This research is funded in part by the National Science Foundation, Award 0927739. Any opinions and/or ndings are those of the author and do not necessarily represent the views of the sponsor. Authors would like to acknowledge the insight and data shared by collaborators Amin Rasekh and Drs. Michael Lindell, Jeryl Mumpower, and Kelly Brumbelow.